\documentclass[conference]{IEEEtran}
\IEEEoverridecommandlockouts
\usepackage{cite}
\usepackage{amsmath,amssymb,amsfonts}
\usepackage{algorithmic}
\usepackage{graphicx}
\usepackage{textcomp}
\usepackage{xcolor}
\def\BibTeX{{\rm B\kern-.05em{\sc i\kern-.025em b}\kern-.08em
    T\kern-.1667em\lower.7ex\hbox{E}\kern-.125emX}}

\usepackage{xspace}

\usepackage{subcaption} 

\makeatletter
\DeclareRobustCommand\onedot{\futurelet\@let@token\@onedot}
\def\@onedot{\ifx\@let@token.\else.\null\fi\xspace}

\def\eg{\emph{e.g}\onedot}

\makeatother

\begin{document}

\title{Edge Detection for Satellite Images without Deep Networks}

\author{\IEEEauthorblockN{Joshua Abraham\IEEEauthorrefmark{1},
Calden Wloka\IEEEauthorrefmark{2},}
\IEEEauthorblockA{Department Electrical Engineering and Computer Science,
York University\\
Toronto, Canada\\
Email: \IEEEauthorrefmark{1}abraham3@my.yorku.ca,
\IEEEauthorrefmark{2}calden@eecs.yorku.ca}}

\maketitle

\begin{abstract}

Satellite imagery is widely used in many application sectors, including agriculture, navigation, and urban planning. Frequently, satellite imagery involves both large numbers of images as well as high pixel counts, making satellite datasets computationally expensive to analyze. Recent approaches to satellite image analysis have largely emphasized deep learning methods. Though extremely powerful, deep learning has some drawbacks, including the requirement of specialized computing hardware and a high reliance on training data. When dealing with large satellite datasets, the cost of both computational resources and training data annotation may be prohibitive.


In this paper, we demonstrate that a carefully designed image pre-processing pipeline allows traditional computer vision techniques to achieve semantic edge detection in satellite imagery that is competitive with deep learning methods at a fraction of the resource costs. We focus on the task of semantic edge detection due to its general-purpose usage in remote sensing, including the detection of natural and man-made borders, coast lines, roads, and buildings.



\end{abstract}

\begin{IEEEkeywords}
edge detection, remote sensing, satellites, image pre-processing
\end{IEEEkeywords}

\section{Introduction}
\label{sec:Introduction}



As satellite imagery continues to become more prevalent in an increasing number of fields, the need to efficiently analyze large volumes of image data has also grown. One challenge, however, is that remote sensing often involves large images, as well as a large number of images, in order to achieve the desired land coverage. As a result, the computational cost of analysis methods is often of great concern.

While deep learning models have seen growing degrees of adoption in remote sensing applications in recent years, offering both high performance on benchmark datasets and end-to-end trainable solutions, there are a number of potential drawbacks. Deep learning methods require specialized hardware to execute efficiently, increasing the cost involved in their application. Additionally, recent work comparing traditional approaches to computer vision with deep learning has shown that while both methods are sensitive to changes in sensor properties used to acquire images, deep learning performance varied much more unpredictably than for classic methods \cite{TsotsosEtAl2019}. Given the large variety of sensor types and sampling conditions used to gather remote sensing data, this is a particular concern for the field \cite{LiEtAl2019}. One potential solution is to develop robust methods based on traditional approaches for common base problems which can be either directly used or integrated into a hybrid deep learning approach \cite{OMahonyEtAl2020}. 

One common problem within the domain of remote sensing is high quality edge extraction, in which semantically meaningful edges are detected while minimizing extraneous edges such as those due to noise. Edge detection is an important precursor in a number of remote sensing applications such as agricultural management \cite{DaiEtAl2020}, ecological monitoring \cite{LiuEtAl2011}, and water body identification \cite{VigneshThyagharajan2017}. 

The Canny edge detector \cite{Canny1986} is a widely used traditional method for edge detection. It is analytically simple to understand, widely implemented in vision software libraries, and computationally efficient to execute. However, it is also vulnerable to noise and is often outperformed by more modern deep learning methods, such as the Holistically-Nested Edge Detection (HED) model \cite{XieTu2015}. However, the advantages that the Canny method provides are particularly well suited to the challenges facing deep learning applications in remote sensing (namely, computational efficiency, wide availability, and analytic simplicity of operation). 

In order to preserve the benefits that a classic method such as Canny provides but improve the quality of edge detection to compete with more modern approaches, we propose the Satellite Pre-processing Enhanced Edge Detection (SPEED) model. SPEED uses a series of pre-processing steps (shown in Figure \ref{fig:sys_model}, see Section \ref{sec:SystemOverview} for full details) to reduce extraneous edge detection and improve the detection and retention of semantically meaningful edges for the Canny edge detector. In Section \ref{sec:Experiments} we compare the performance of SPEED to the unmodified Canny detector and a deep learning based edge detection method, HED \cite{XieTu2015}, on the detection of ground-truth labeled edges in the DOTA dataset \cite{XiaEtAl2018}. SPEED dramatically improves the performance of the Canny method with only a modest increase in computational cost, outperforming HED in both detection properties and computational efficiency.

\subsection{Related Work}


A number of prior works have used pre-processing filters to reduce issues of noise, cloud coverage, shadows, and incident light variations within satellite imagery. Starovoitov and Makarau \cite{StarovoitovMakarau2008} created a pre-processing pipeline consisting of Kuwahara filters, multi-resolution image fusion, and a Laplacian filter for the classification of multi-spectral images. Similarly, Liu \& Jezek \cite{LiuJezek2004} apply a combination of Gaussian filtering and anisotropic diffusion to satellite images as a pre-processing stage of a processing pipeline specifically focused on coastline extraction. Though the goals of these systems differs from ours (namely, classification and coastline extraction, as opposed to general semantic edge detection), the application of a series of back-to-back filters designed to handle some of the specific challenges and patterns of noise present in satellite imagery provided useful guidance when designing our own system. 

More recently, Krauß \cite{Krauss2015} focused on semantic edge extraction through the construction of a depth map (or digital surface model) of the satellite image. Krauß employs a series of pre-processing steps to improve both the depth construction as well as the resultant edge extraction. While effective, this method requires specialized sources of data including stereoscopic images and ground sensor data, limiting its generality for widespread adoption.

\section{System Overview}
\label{sec:SystemOverview}

\begin{figure}
\centerline{\includegraphics[width=\linewidth]{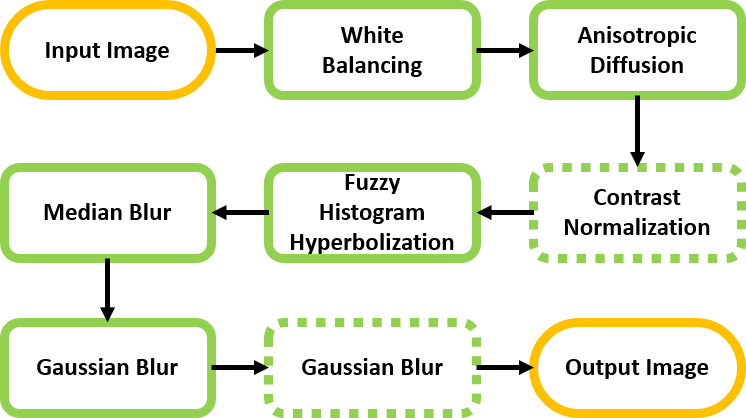}}
\caption{Steps in the Satellite Pre-processing Enhanced Edge Detection (SPEED) pipeline. Conditional steps are shown with dotted borders.}
\label{fig:sys_model}
\end{figure}

SPEED provides a general method for semantic edge detection designed to operate over monocular input images without the need for additional data streams. SPEED makes use of seven pre-processing steps, combining techniques meant to remove many of the common issues associated with remote sensing as with steps for edge detection enhancement. Figure \ref{fig:sys_model} provides the processing stream, where each step is successively applied to the satellite image. Within this algorithm, there are two steps that are conditionally applied if specific conditions within the image are met; these are indicated with a dashed box in Figure \ref{fig:sys_model}. These conditional steps improve detection quality for specific remote sensing challenges present on only a subset of input images, such as images with very high anisotropy in brightness levels (\eg due to reflections off water along a coast). 

After pre-processing, images are sent through a Canny edge detector. This edge detector was chosen because it is computationally inexpensive, widely available, well understood, semantically interpretable, and does not does rely on extensive training data.  

\subsection{White Balancing}
\label{subsec:WB}

The first step within the pipeline is white balancing. This step must be done first within the pipeline as it is a colour dependant step; following white balancing, the image is converted to grayscale for the remainder of the SPEED pipeline. 

The purpose of white balancing is to remove any coloured haze on the image to make it look as if it is under white light. Haze on an image can impede the accuracy of edge detection algorithms by reducing local contrast, possibly resulting in edges being missed or incorrectly detected. This issue becomes more prevalent for satellite images when there are varying incident angles for the light, or may be exacerbated by weather and atmospheric variances. 

\begin{figure}
    \begin{subfigure}[b]{0.48\linewidth}
        \centerline{\includegraphics[width=\linewidth]{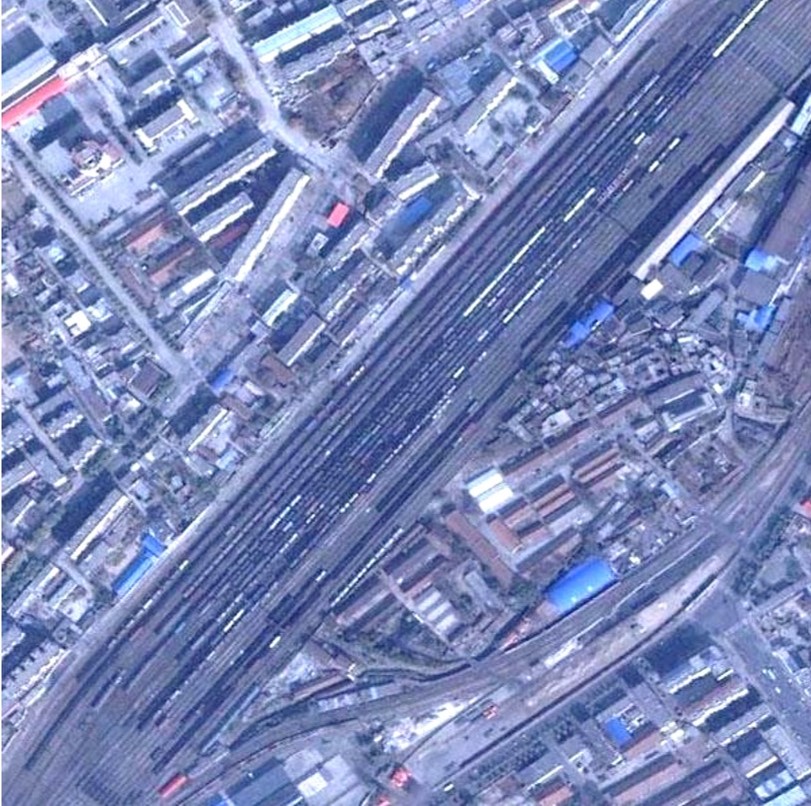}}
        \caption{Original image \label{subfig:wb_orig}}
    \end{subfigure}
    \begin{subfigure}[b]{0.48\linewidth}
        \centerline{\includegraphics[width=\linewidth]{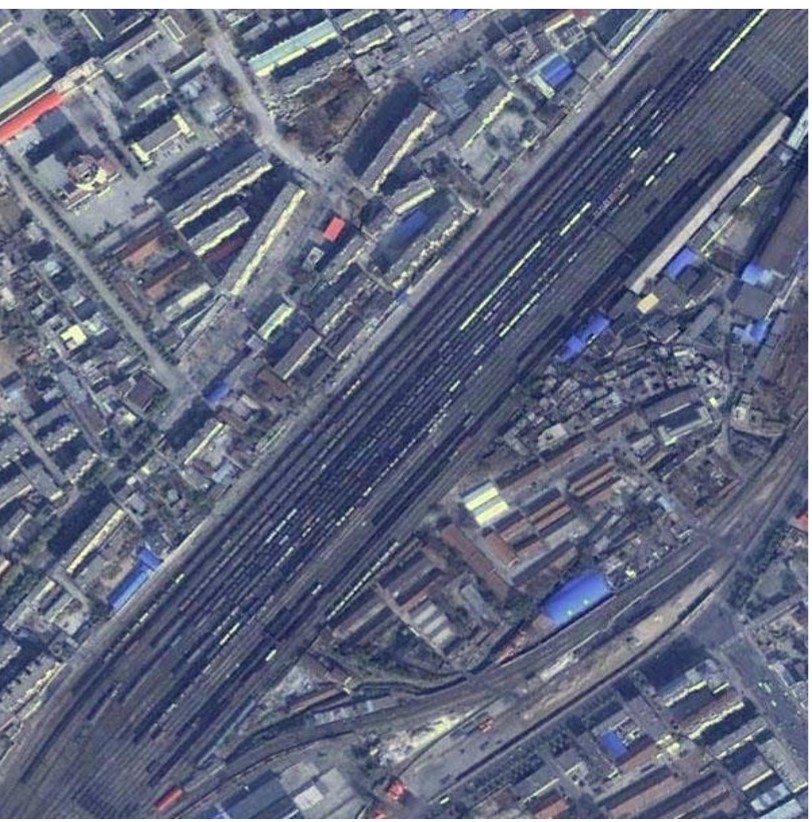}}
        \caption{After white balancing \label{subfig:wb_applied}}
    \end{subfigure}
\caption{Example image showing the effect of white balancing.}
\label{fig:white_balance}
\end{figure}

Due to the highly variable lighting conditions under which satellite imagery is gathered, this is an extremely impactful step on the resulting quality of edge detection. For example, Figure \ref{fig:white_balance} shows a satellite image of a train track adversely affected by hazing (Figure \ref{subfig:wb_orig}). Following the application of white balancing (Figure \ref{subfig:wb_applied}), the image has greatly improved contrast, most apparent in the distinctive edges seen on lighter coloured buildings and trains.


\subsection{Anisotropic Diffusion}
\label{subsec:AD}

Anisotropic diffusion is a technique used to maintain semantically meaningful parts of an image while reducing the overall noise. Anisotropic diffusion (sometimes referred to as Perona-Malik Diffusion) uses a stochastic differential equation to create a group of increasingly blurred images over an input image, $I$, according to the space invariant differential equation defined in Equation \ref{eq:Anisotropic}, where $\nabla$ represents the gradient, and $\nabla^2$ represents the Laplacian. The variable $c$ within Equation \ref{eq:Anisotropic} is known as the \emph{diffusion coefficient} and is used to determine the rate of diffusion. The original paper defined two functions based on the gradients of the image to determine the diffusion constant. The first function, given in Equation \ref{eq:Coefficient}, was found to privilege high contrast images and was empirically tested to produce better results within SPEED. The $K$ value can be defined either by empirical testing or, as is used in the results reported in this paper, be set to the output of the Canny noise estimator \cite{Canny1986}. 

\begin{equation}
    \label{eq:Anisotropic}
    I' =\nabla c * \nabla I + c * \nabla^2 I
\end{equation}
\begin{equation}
    \label{eq:Coefficient}
    c = e^{-\left( \frac{||\nabla I ||}{K} \right)^2}
\end{equation}

This step must be done prior to the conditional contrast normalization (Section \ref{subsec:CN}) as the normalization will result in smaller contrast variances. If done first, the contrast normalization filter would reduce the edges accentuated within this step. This step focuses on high contrast edges, and the effect of this filter is later strengthened by Fuzzy Histogram Hyperbolization (Section \ref{subsec:FHH}) which operates over perceived brightness and local contrast to improve the capture of low contrast edges.

\begin{figure}
  \begin{center}
    \begin{subfigure}[b]{0.48\linewidth}
        \centerline{\includegraphics[width=\linewidth]{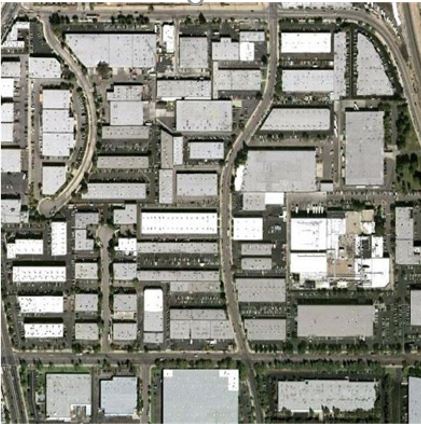}}
        \caption{Original image \label{subfig:anisotropic_a}}
    \end{subfigure}
    \\
     \begin{subfigure}[b]{0.48\linewidth}
        \centerline{\includegraphics[width=\linewidth]{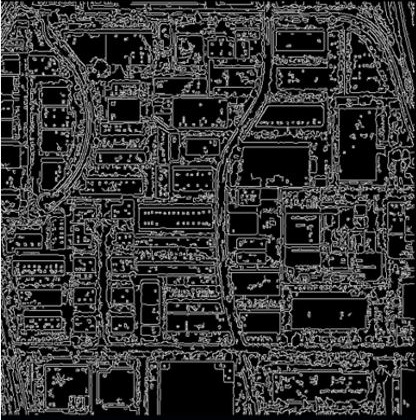}}
        \caption{Canny output without anisotropic filtering \label{subfig:anisotropic_b}}
    \end{subfigure}
    \begin{subfigure}[b]{0.48\linewidth}
        \centerline{\includegraphics[width=\linewidth]{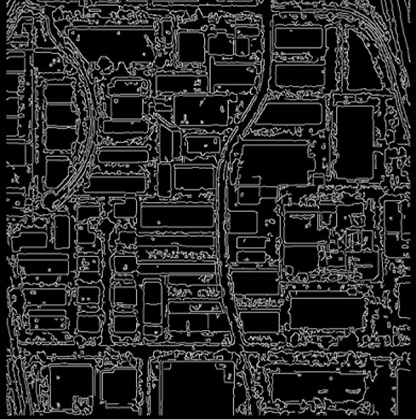}}
        \caption{Canny output with anisotropic filtering \label{subfig:anisotropic_c}}
    \end{subfigure}
    \end{center}
\caption{Example image showing the effect of the anisotropic filter, which allows for significant structural edges within an image to be maintained while extraneous edges are removed.}
\label{fig:anisotropic}
\end{figure}

Figure \ref{fig:anisotropic} provides an example image demonstrating the usefulness of this step. Figures \ref{subfig:anisotropic_b} and \ref{subfig:anisotropic_c} demonstrates the Canny output of the satellite image without and with the addition of the anisotropic filter, respectively. The anisotropic filter helps maintain structurally significant aspects of the image while reducing noisy edge detection. For example, the winding road within the image has many significant edges that are more clearly outlined in the processed image.

\subsection{Conditional Dual Sided Contrast Normalization}
\label{subsec:CN}

Satellite imagery may contain highly reflective surfaces, such as bodies of water or glass buildings, which leads to some images having extremely bright or dark portions of the image. This step aims to identify images in which this is likely to have occurred and re-normalize the image to reduce contrast skew. 

This step in SPEED is only run if a high level of skew is detected in the distribution of pixel values. Skew is determined by constructing a histogram of image pixel values and taking the ratio of the uppermost bins against the lowermost bins to see if the image is skewed towards one of these directions. If there is skew in one of these directions, then the values will be shifted upwards or downwards by a factor of 20\%, and then re-normalized to ensure no saturation occurs. This step is done prior to blurring (Section \ref{subsec:GB}) to ensure that additional contrast skew is not artificially generated from the spread of a single region that is truly higher contrast.

\subsection{Fuzzy Histogram Hyperbolization}
\label{subsec:FHH}

Histogram Hyperbolization is a non-linear transformation that enhances local variations of brightness within an image and creates images that better represent human perception of brightness \cite{Frei1977}. This approach differs from Histogram Equalization, which aims to equalize brightness values across an image according to their histogram results \cite{PatelEtAl2013}. Instead, within the hyperbolic approach, brightness values are altered based on perceived brightness rather than absolute brightness. Tizhoosh \& Fochem \cite{TizhooshFochem1995} introduced the Fuzzy Histogram Hyperbolization (FHH) variant, which we adopt in SPEED. 

FHH allows for more consistent processing results across a wide variety of types of images, which is essential in satellite imagery due to the high degree of data variance. FHH uses a logarithmic scale when altering the brightness within an image on a per-pixel basis, which more closely resembles how humans perceive light. This allows for localized contrast to become more significant within an image, allowing for the capture of additional edge components that may have previously been missed. 

The following calculations are applied for each pixel within the image where $g_{ij}$ represents the intensity of the pixel in the $i$th column, and $j$th row. Within the SPEED pipeline, the standard membership function, Equation \ref{eq:membershipFHH} defined within the original paper is used; where $g_{max}$ and $g_{min}$ respectively represent the global highest and lowest brightness intensities within the image. Once the normalized values are computed, they are then scaled by Equation \ref{eq:lambdaFHH}, where $L$ represents the chosen number of histogram buckets. 
\begin{equation}
    \label{eq:membershipFHH}
    \mu (g_{ij}) = \frac{g_{ij}-g_{min}}{g_{max}-g_{min}}
\end{equation}

\begin{equation}
\label{eq:lambdaFHH}
    \lambda = \frac{L-1}{e^{-1}-1}
\end{equation}
The $\lambda$ scaling function is then multiplied by the previously defined normalization function, and the final pixel values are given by:
\begin{equation}
    \label{eq:finalFHH}
    g_{ij} = \left( \frac{L-1}{e^{-1}-1} \right)*(e^{\mu(g_{ij})}-1)
\end{equation}

\begin{figure}
    \begin{subfigure}[b]{0.48\linewidth}
        \centerline{\includegraphics[width=\linewidth]{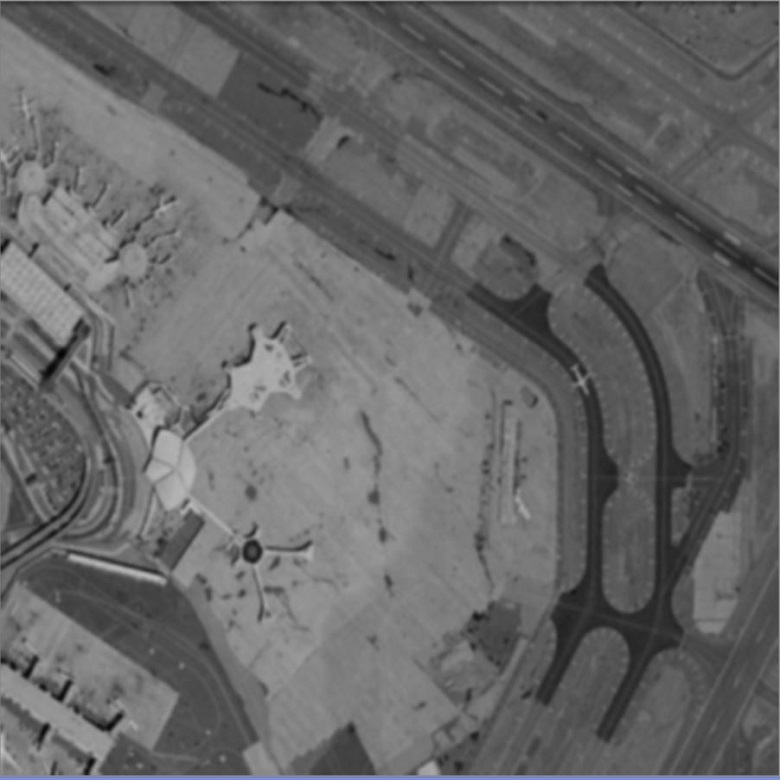}}
        \caption{Image without FHH \label{subfig:no_fhh_a}}
    \end{subfigure}
    \begin{subfigure}[b]{0.48\linewidth}
        \centerline{\includegraphics[width=\linewidth]{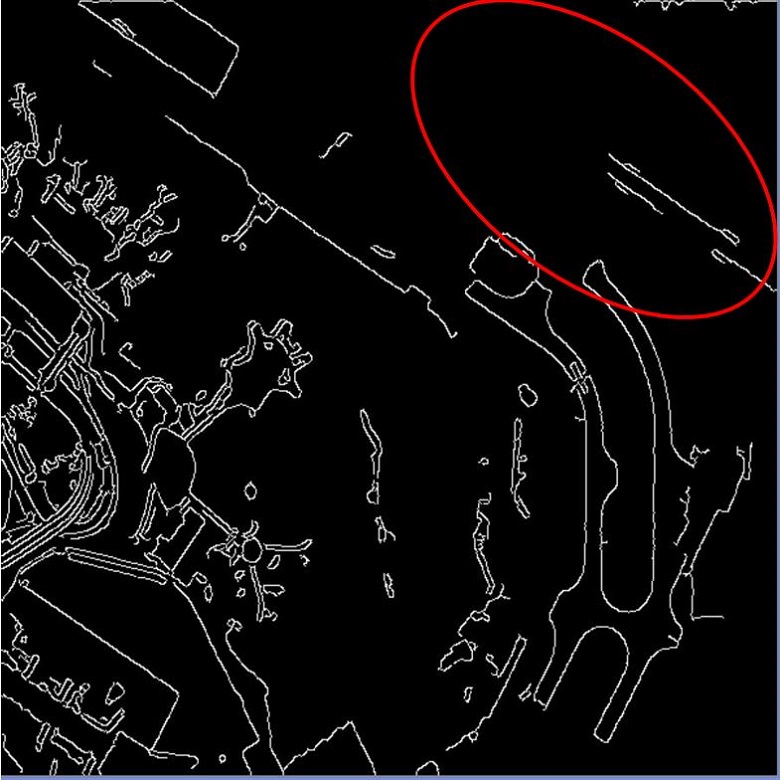}}
        \caption{Canny output without FHH \label{subfig:no_fhh_b}}
    \end{subfigure} \\ \vspace{-1mm} \\
     \begin{subfigure}[b]{0.48\linewidth}
        \centerline{\includegraphics[width=\linewidth]{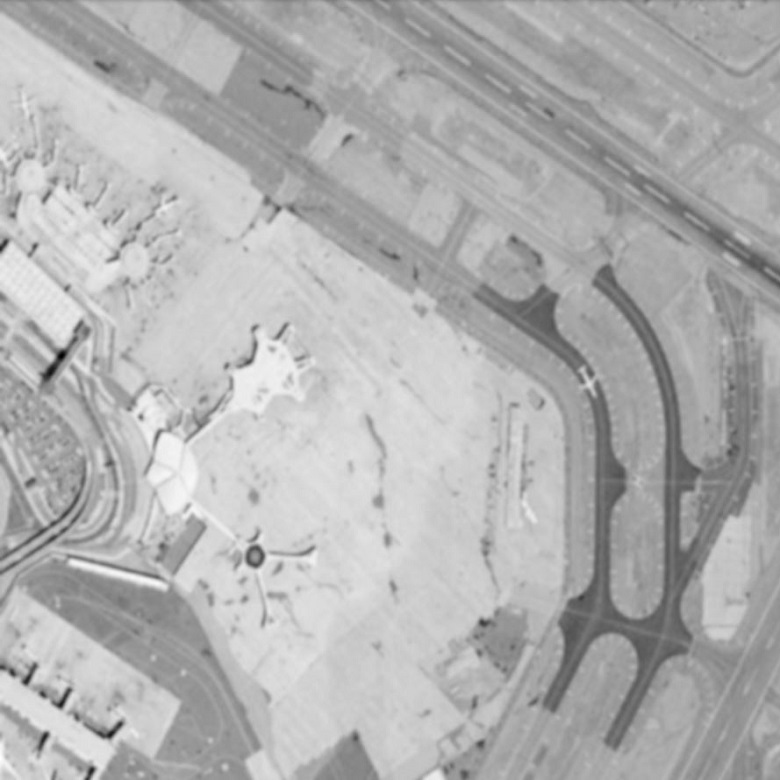}}
        \caption{Image with FHH \label{subfig:fhh_a}}
    \end{subfigure}
    \begin{subfigure}[b]{0.48\linewidth}
        \centerline{\includegraphics[width=\linewidth]{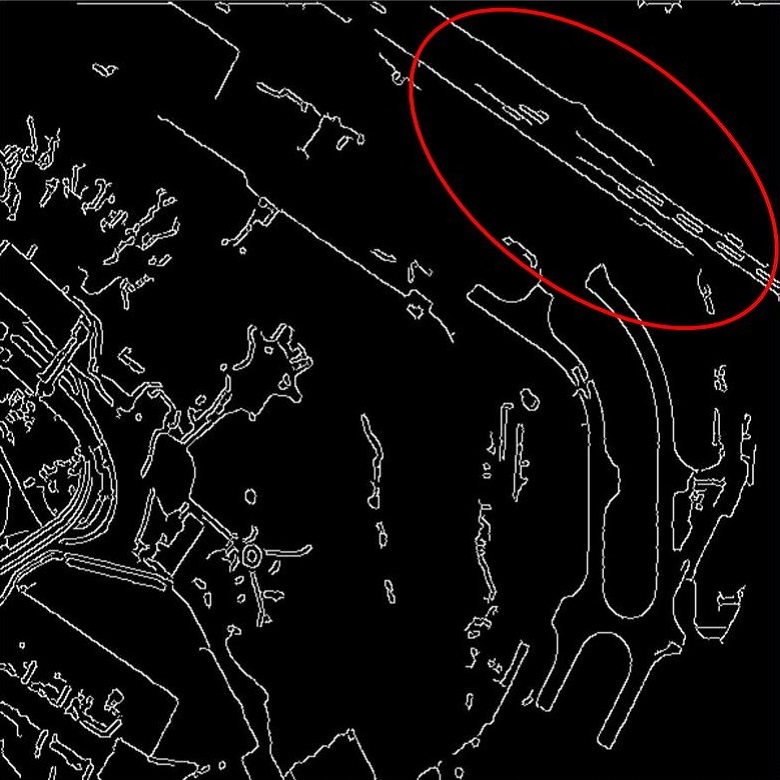}}
        \caption{Canny output with FHH \label{subfig:fhh_b}}
    \end{subfigure}
\caption{Example images showing the effect of FHH. Note that within the red highlighted region a previously poorly detected road segment is now successfully found following the application of FHH.}
\label{fig:fhh}
\end{figure}

In Figure \ref{fig:fhh}, the regions circled in red in Figures \ref{subfig:no_fhh_b} and \ref{subfig:fhh_b} highlight the importance of applying FHH as the localized contrast in that region is accentuated following FHH, allowing the previously undetected road contour to be successfully found. 

\subsection{Median Blur}

Current state-of-the-art satellites are often not plagued by salt and pepper noise, but using a median blur filter still provides some benefit. Small details or variances within an image at high altitudes are typically not significant when looking for edges. These small variances represent minute details like cars on a road, trees in a field, or windows on a building, and do not pose a benefit to keep when performing edge detection. These small variances add many extraneous edge elements to the image and may detract from major structural edges which are present. Although helpful for removing small extraneous elements, median blurring is a destructive filtering step, and so we found the best results were found when the filter was kept small (a $3\times3$ median blur kernel). 

\begin{figure}
  \begin{center}
    \begin{subfigure}[b]{0.48\linewidth}
        \centerline{\includegraphics[width=\linewidth]{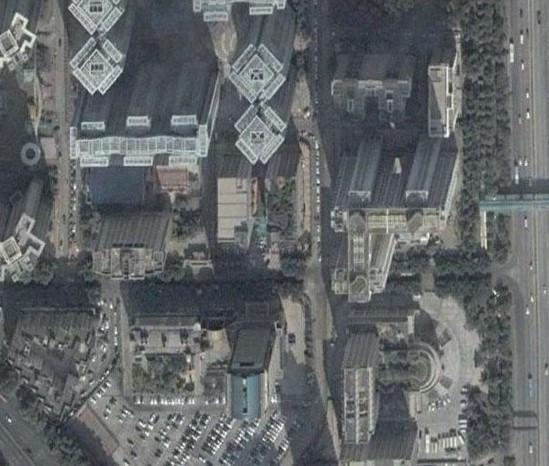}}
        \caption{Original image \label{subfig:median_orig}}
    \end{subfigure}
    \end{center}
     \begin{subfigure}[b]{0.48\linewidth}
        \centerline{\includegraphics[width=\linewidth]{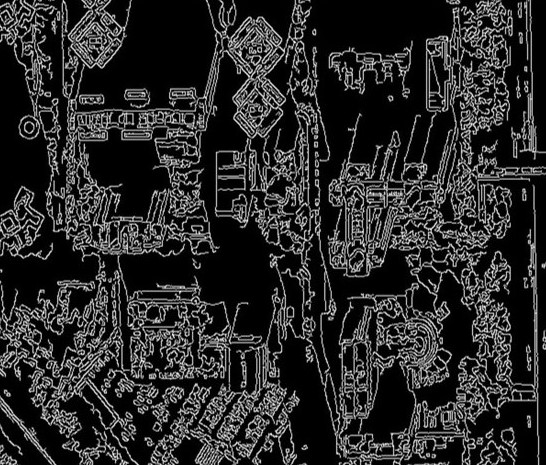}}
        \caption{Canny output without median blurring applied \label{subfig:no_median_canny}}
    \end{subfigure}
    \begin{subfigure}[b]{0.48\linewidth}
        \centerline{\includegraphics[width=\linewidth]{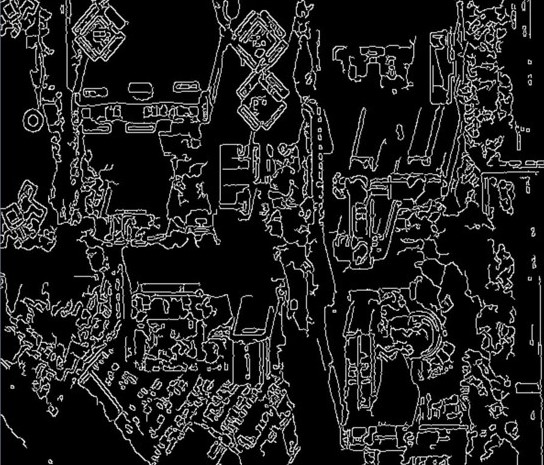}}
        \caption{Canny output with median blurring applied \label{subfig:median_canny}}
    \end{subfigure}
\caption{Example image showing the effect of median blurring when using a 3x3 median blur kernel.}
\label{fig:median_blur}
\end{figure}

Figure \ref{fig:median_blur} shows an example of the effect of median blurring. Figure \ref{subfig:no_median_canny} shows the output of the Canny detector without median filtering, while \ref{subfig:median_canny} shows the Canny output with a $3\times3$ median blur kernel. Note that following median blurring the number of edge elements found is greatly reduced without strongly affecting the detection of large, semantically meaningful edge detection.

\subsection{Gaussian Blur}
\label{subsec:GB}

Gaussian blurring (sometimes referred to as \emph{smoothing}) an image reduces minor variations within the overall image by spreading values over a small neighbourhood. In SPEED this step happens later in the pipeline, as a number of the prior filters operate over small local neighbourhoods in order to reduce various types of noise or accentuate edge contrast within the image. Blurring prior to these steps would make these regions more uniform, and thus potentially disrupt the normal operation of those filters. The use of a small Gaussian kernel was found to be most effective, so a $3\times3$ Gaussian kernel was used with a $\sigma$ value of 1.5.

\subsection{Conditional Gaussian Blur}

When designing the SPEED model, we found that the optimal degree of Gaussian blurring is not uniform across different images. While the kernel described in Section \ref{subsec:GB} provided a sufficient degree of smoothing without hampering the detection of significant edge structures, some images were improved by a stronger degree of blurring. This can be accomplished by applying a second Gaussian blur (equivalent to the application of a single larger blurring kernel).

In order to determine whether a stronger degree of blurring is required, we want to determine if the image still contains a large number of small variances which may lead to extraneous edges found by the Canny detector. This is done by constructing the image histogram following the Gaussian blur described in Section \ref{subsec:GB} with one bin for each possible pixel value. Based on this histogram, a calculation is done to determine how many non-empty buckets contain at most 0.01\% of pixels. If more than 10\% of the bins satisfy this property, then the secondary Gaussian blur will be run. In the default configuration of SPEED, this second Gaussian blur uses the same kernel parameters as described in Section \ref{subsec:GB}.

\section{Experiment Results}
\label{sec:Experiments}

In order to evaluate the behaviour of SPEED, we use the DOTA dataset \cite{XiaEtAl2018}. DOTA contains 2806 aerial images of varying sizes, as well as ground truth semantic labels for significant objects of interest. We extract the boundaries of all ground truth labeled targets to construct ground-truth edge images. Note that not all semantically meaningful edges are included in the DOTA list of targets, but the density and variety of target labels nevertheless provides a good baseline for performance evaluation.

In order to compute accuracy, we make use of a modified evaluation based on the Structural Similarity Index Measure (SSIM). We first compute a pixel-wise SSIM map using a Gaussian kernel window with $\sigma=1.5$ comparing the ground truth edge image to a predicted edge map. The SSIM map is then binarized, such that all pixels with SSIM score greater than $0$ are set to $1$, producing a \emph{base matching map} ($M$). The binarization of the SSIM map ensures that predicted edges do not need to exactly coincide with ground truth edges in order to count as found, but rather simply need to be close enough to garner a positive SSIM score. $M$ therefore represents all pixels which share structural similarity between the ground truth edge map and the predicted edge map.

To compute the true positive rate, $TP_{score}$, we employ the following computation:

\begin{equation}
    \label{eq:truePositive}
     TP_{score}= \dfrac{\sum_{j=0}^{y}  \sum_{i=0}^{x} M_{ij} \land G_{ij}}{\sum_{j=0}^{y}  \sum_{i=0}^{x} G_{ij}}
\end{equation}
where $G$ is the ground truth edge map and subscript $ij$ refers to the map value at coordinates $(i,j)$. This equation finds the sum of all ground truth pixels with detected structural similarity in the predicted edge map.

Similarly, we compute a false positive score, $FP_{score}$, according to the equation:

\begin{equation}
    \label{eq:falsePositive}
     FP_{score}= \dfrac{\sum_{j=0}^{y}  \sum_{i=0}^{x} \neg M_{ij} \land D_{ij}}{\sum_{j=0}^{y}  \sum_{i=0}^{x} G_{ij}}
\end{equation}
where $D$ is the predicted edge map of the edge detector method being evaluated. This equation therefore finds the total number of predicted edge pixels which do have any match with the structure of the ground truth map, normalized by the total number of ground truth edge pixels. Note that this false positive score is not bounded between $0$ and $1$, though in practice we find the scores tend to fall within that range.

\begin{figure}
  \begin{center}
    \begin{subfigure}[b]{0.48\linewidth}
        \centerline{\includegraphics[width=\linewidth]{original_noHED.jpg}}
        \caption{Original image \label{subfig:orig_noHED}}
    \end{subfigure}
    \end{center}
     \begin{subfigure}[b]{0.48\linewidth}
        \centerline{\includegraphics[width=\linewidth]{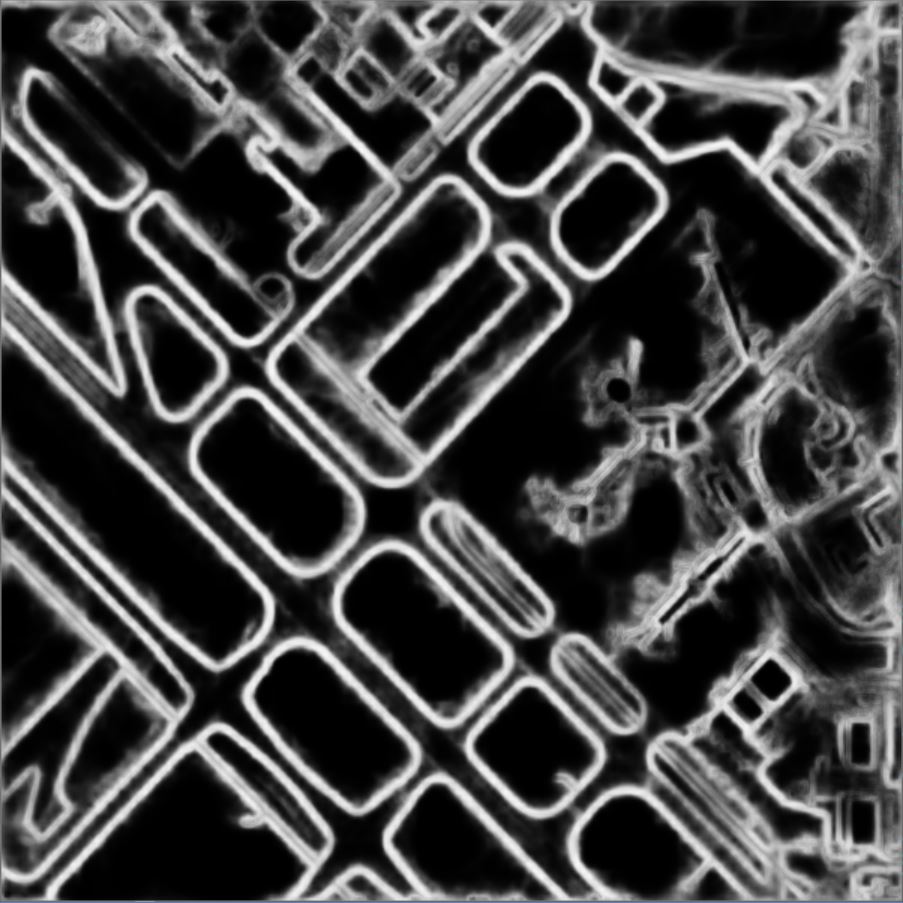}}
        \caption{HED output \label{subfig:hed_output}}
    \end{subfigure}
    \begin{subfigure}[b]{0.48\linewidth}
        \centerline{\includegraphics[width=\linewidth]{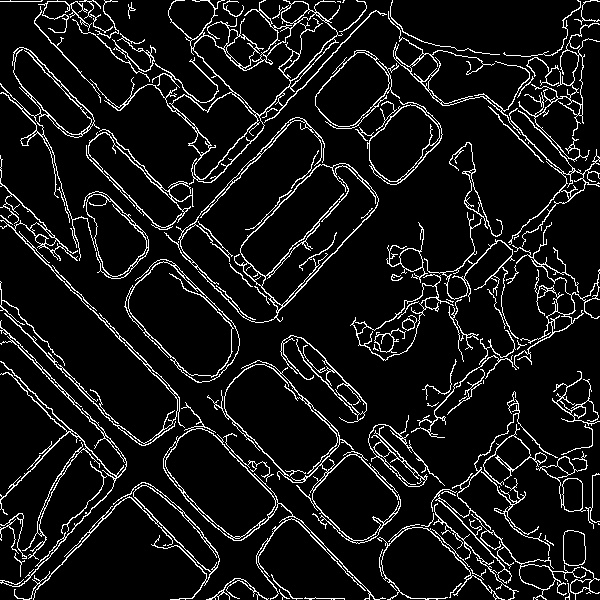}}
        \caption{Canny HED output \label{subfig:hed_canny_output}}
    \end{subfigure}
\caption{Example image showing the HED output in a Canny-like format}
\label{fig:hed_to_canny}
\end{figure}

We compare the output of SPEED with the raw Canny edge detector and the HED \cite{XieTu2015} model. However, in order to fairly evaluate the HED model, we needed to slightly modify its output to match the format of both the ground truth maps and the maps produced by the Canny edge detector. In its original formulation, HED marks detected edges with thick lines more than one pixel thick (see Figure \ref{subfig:hed_output}). Since the ground truth maps do not contain thick lines, this inadvertently penalized the HED method. We therefore converted the HED output to a Canny-like map by running a Canny edge detector on the HED map (Figure \ref{subfig:hed_canny_output}). This retains the salient information of the original HED output, but ensures that the numerical comparison between HED and SPEED is done in an equivalent manner. Subsequent results and examples given for the HED method, therefore, refer to the HED method with output modified in this manner.

In addition to evaluating detection accuracy (Section \ref{subsec:DetectionPerformance}), we also compare SPEED and HED based on computation time (Section \ref{subsec:TimePerformance}). Section \ref{subsec:SPEED_configuration} further explores the design of the SPEED method, including ablation studies and the effect of re-configuring the filter order.

\subsection{Edge Detection Performance}
\label{subsec:DetectionPerformance}

\begin{figure}
\centerline{\includegraphics[width=\linewidth]{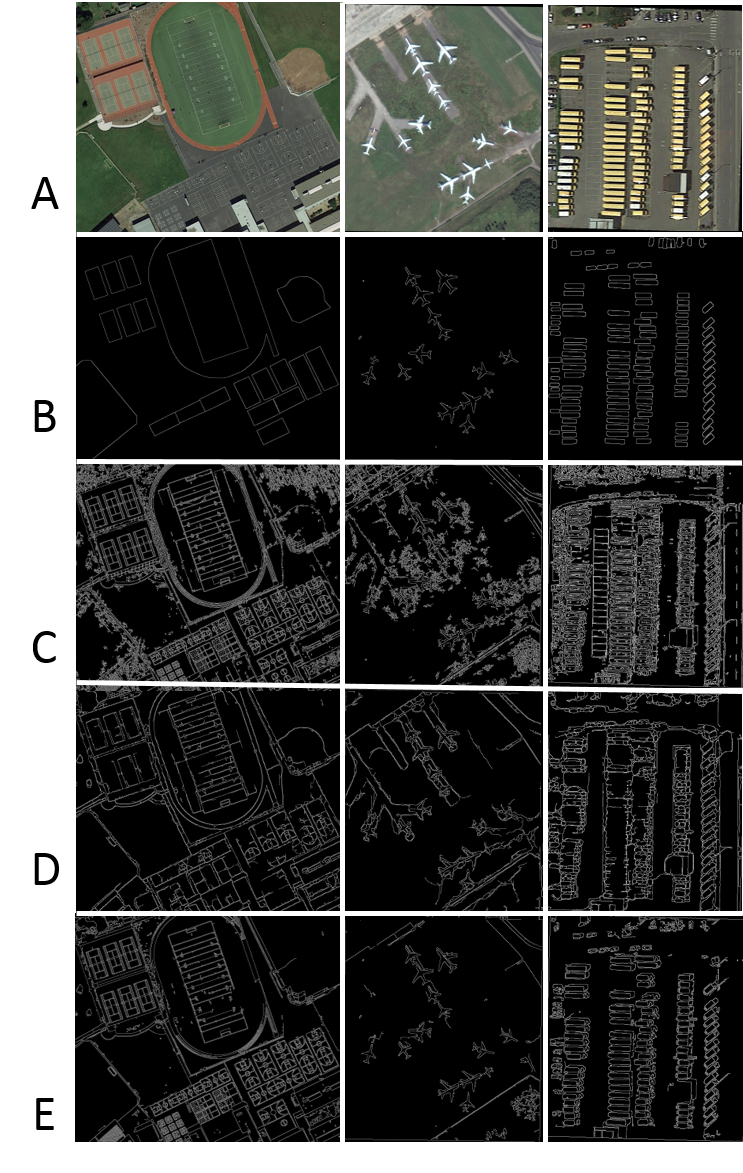}}
\caption{Example edge predictions. A: original image. B: ground truth. C: Canny raw output. D: HED. E: SPEED (our method).}
\label{fig:image_stitch}
\end{figure}

Example output of the methods being compared on three images from the DOTA dataset is shown in Figure \ref{fig:image_stitch}. Note that when compared to the raw Canny output, SPEED drastically reduces extraneous edge detections while maintaining and improving the detection of semantically important edges.

\begin{figure}
\centerline{\includegraphics[width=0.8\linewidth]{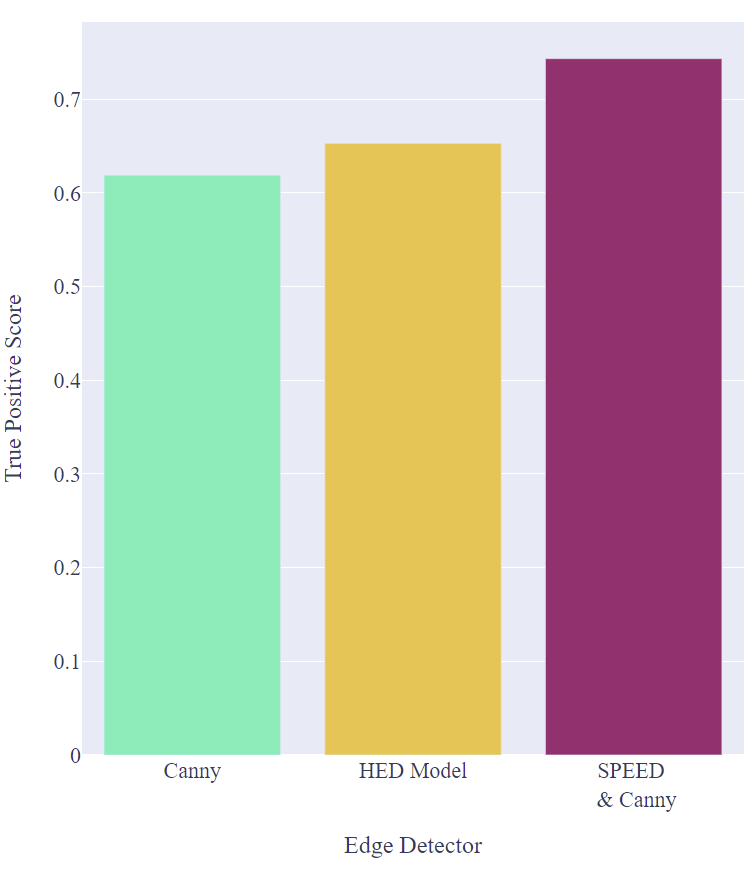}}
\caption{True Positive ratios for the raw Canny, HED model, and SPEED. Higher scores mean better performance.}
\label{fig:edgeDetectors_positive}
\end{figure}

\begin{figure}
\centerline{\includegraphics[width=0.8\linewidth]{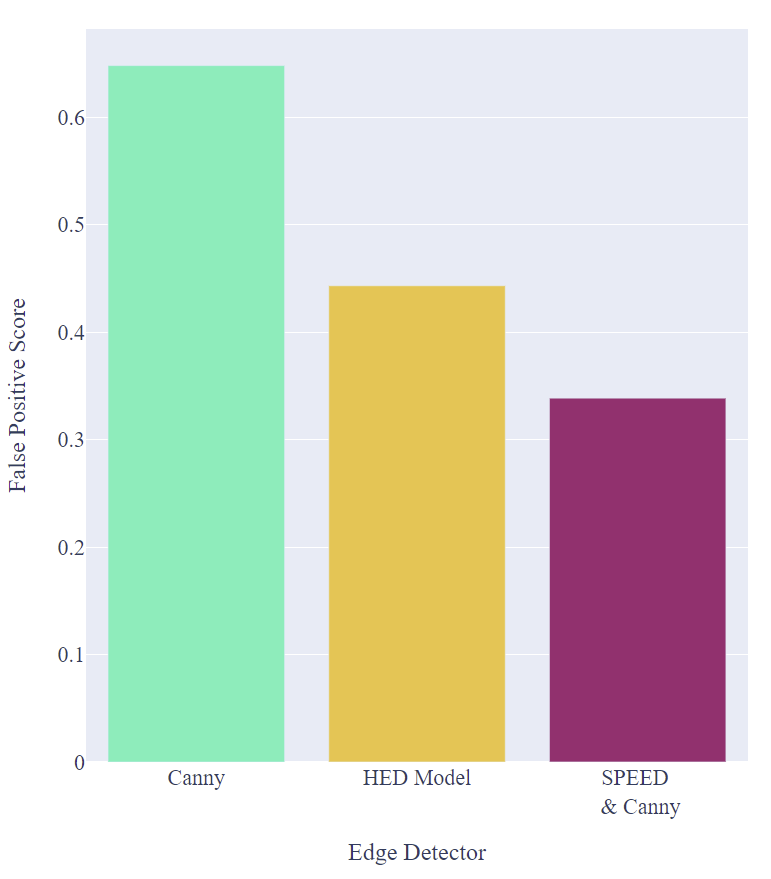}}
\caption{False Positive scores for the raw Canny, HED model, and SPEED. Lower scores mean better performance.}
\label{fig:edgeDetectors_negative}
\end{figure}

Detection accuracy is evaluated as the combination of both the true positive rate (Figure \ref{fig:edgeDetectors_positive}) and false positive score (Figure \ref{fig:edgeDetectors_negative}). It can be seen that SPEED not only dramatically outperforms raw Canny, it also has both a higher true positive rate and a lower false positive score than the HED model.

\subsection{Computation Time}
\label{subsec:TimePerformance}

As previously mentioned, remote sensing often involves very large quantities of imaging data to analyze. The computational efficiency of methods employed to perform that analysis, therefore, is also an important factor. Most deep learning models are designed to run on a system with specialized hardware, such as a graphical processing unit (GPU) or tensor processing unit (TPU). This introduces an additional cost over more traditional computer processor unit (CPU) only systems, including both hardware costs as well as greater power requirements.

\begin{figure}
\centerline{\includegraphics[width=0.8\linewidth]{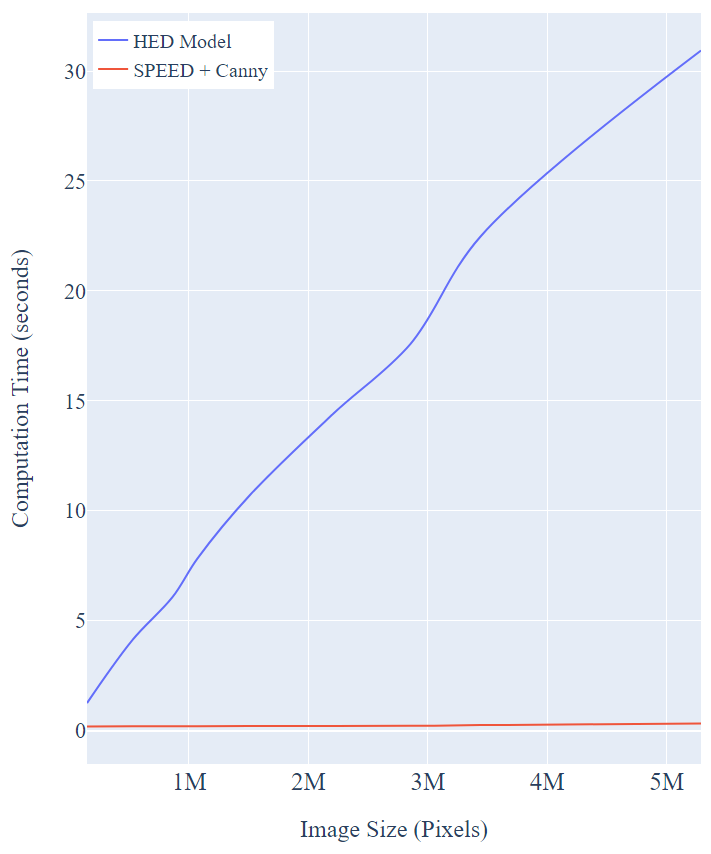}}
\caption{Computation time comparisons between the HED model, and the SPEED pipeline with Canny for various images sizes}
\label{fig:timeAnalysis}
\end{figure}

We demonstrate the computational advantage that SPEED provides when restricted to a CPU only system. Figure \ref{fig:timeAnalysis} shows the relationship between computation time required and image size for both the HED model and the SPEED pipeline. Note that the computational time required to execute SPEED increases relatively slowly, whereas the requirements for HED increase dramatically with increasing image size, rapidly reaching execution time which is prohibitive for large datasets. In contrast, the computational resource requirements for SPEED are greatly reduced, making it a feasible solution for a wider variety of remote sensing applications.

\subsection{SPEED Configuration Analysis}
\label{subsec:SPEED_configuration}


\begin{table}[]
    \centering
    \begin{tabular}{p{0.13\linewidth} | p{0.13\linewidth} | p{0.13\linewidth} | p{0.20\linewidth} | p{0.20\linewidth}}
    Conditional Filter & $TP_{score}$ No Filter & $FP_{score}$ No Filter & $TP_{score}$ Always On & $FP_{score}$ Always On  \\
     \hline
    
    CB & 0.728 & 0.347 & 0.729 & 0.344   \\
    CN & 0.735 & 0.367 & 0.687 & 0.412   \\
    \end{tabular}
    \caption{Accuracy scores showing the effect of removing the conditional steps or switching them to being always applied. Filters are given using the following abbreviations: \\ \emph{CN}~-~Contrast Normalization, \emph{CB} - Conditional (Gaussian) Blur.  \label{tab:conditional_results}}
\end{table}

This section explores the behaviour of the SPEED pipeline in more detail through a set of ablation studies and configuration adjustments. For the ablation studies, the pipeline was tested after removing a single component in order to assess the overall impact of that component. The steps examined in this way include: secondary conditional blurring, conditional contrasting, Fuzzy Histogram Hyperbolization, and anisotropic diffusion. We focused on these components because they are comparatively less common that the other pipeline steps, whose benefits are more well understood.

\begin{figure}
\centerline{\includegraphics[width=0.9\linewidth]{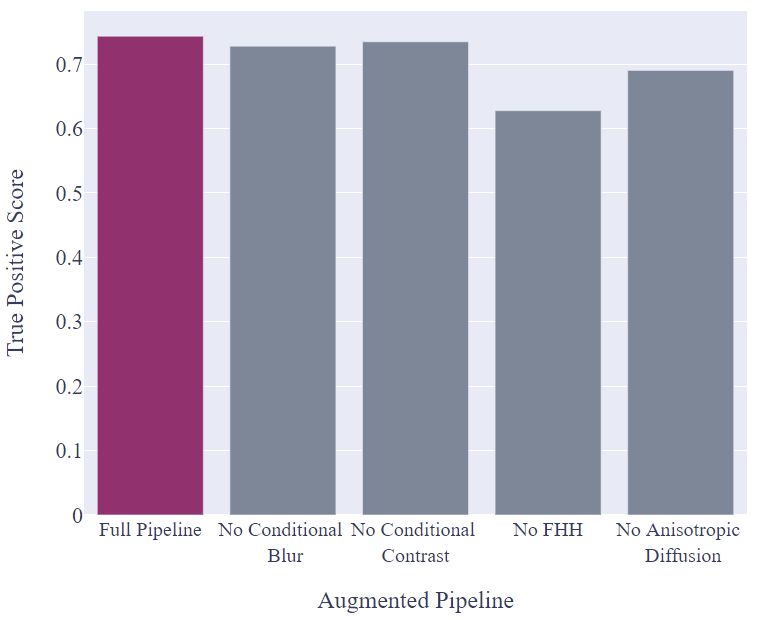}}
\caption{True positive rate following the removal of a pipeline component. The full pipeline performance is given on the left in colour for comparison.}
\label{fig:ablationTrue}
\end{figure}

\begin{figure}
\centerline{\includegraphics[width=0.9\linewidth]{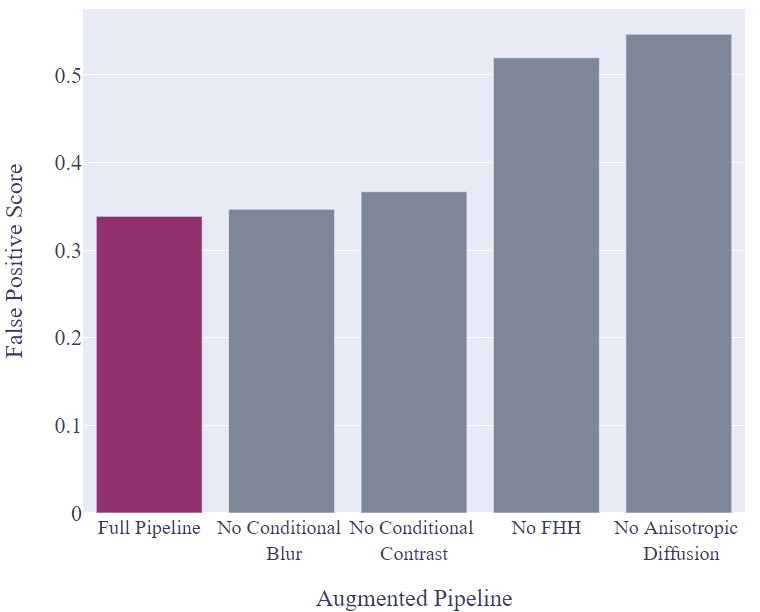}}
\caption{False positive score following the removal of a pipeline component. The full pipeline performance is given on the left in colour for comparison.}
\label{fig:ablationFalse}
\end{figure}

The results of this analysis are shown in Figures \ref{fig:ablationTrue} and \ref{fig:ablationFalse}, with the original SPEED pipeline highlighted for comparison. The removal of the conditional steps has some affect on performance, but they are notably less impactful than the other components tested due to only being invoked on a sub-portion of the test set. Over the 2806 images in the DOTA dataset, only 893 images met the conditions for a secondary blur, and only 640 met the conditions for contrast normalization. We additionally tested the effect of always invoking these conditional steps, the results of which are shown in Table \ref{tab:conditional_results}. As can be seen, always running the conditional filters is either similar to never running them, in the case of the conditional blur, or more detrimental, in the case of the conditional contrast normalization. Thus, while these filters do conditionally provide a benefit and are important for handling a challenging subset of input images, they should not be turned on by default. 

The removal of the Fuzzy Histogram Hyperbolization had the most notable impact on the true positive score. When removed the filter results in a decrease in capture of 28.77\%. Finally, the anisotropic diffusion filter provides the most significant improvement to the reduction of semantically unimportant edges, indicating the importance of noise suppression for Canny.

\begin{table}[]
    \centering
    \begin{tabular}{ccc}
    SPEED Filter Order  & $TP_{score}$  & $FP_{score}$ \\
     \hline
    \emph{WB-AD-CN-FHH-MB-GB-CB} & \emph{0.743} &\emph{0.339}  \\
    WB-AD-MB-GB-CB-FHH-CN & 0.617 & 0.557     \\
    WB-CN-FHH-GB-CB-MB-AD & 0.680 & 0.436     \\
    WB-FHH-AD-MB-GB-CB-CN & 0.577 & 0.645     \\
    WB-MB-GB-CB-CN-FHH-AD & 0.724 & 0.350     
    \end{tabular}
    \caption{SSIM based $True_{positive}$ and $False_{positive}$ scores for different SPEED configurations. Filter orders are given using the following abbreviations: \\ \emph{WB}~-~White Balancing, \emph{AD}~-~Anisotropic Diffusion, \\ \emph{CN}~-~Contrast Normalization, \emph{FHH}~-~Fuzzy Histogram \\ Hyperbolization, \emph{MB} - Median Blur, \emph{GB} - Gaussian Blur, \\ \emph{CB} - Conditional (Gaussian) Blur. Results for the standard pipeline are shown in italics. \label{tab:order_results}}
\end{table}

In addition to the removal or addition of steps in the pipeline, we also experimented with reordering the filtering components in order to examine the impact of the specific internal ordering. As the white balancing step must be run on colour images, it was maintained as the primary filter and remained the only stationary component. The results of this experiment are summarized in Table \ref{tab:order_results}.

\section{Conclusion}

In this paper, we present the SPEED method for semantic edge detection in satellite imagery. This method retains many of the advantages of the Canny edge detector, including computational efficiency, while still outperforming comparable deep learning methods.


\section*{Acknowledgment}

We would like to thank the York University Lassonde School Research Support Team for their financial support.

{\small
\bibliographystyle{IEEEtran}
\bibliography{crv_2021}
}

\end{document}